# StonkBERT: Can Language Models Predict Medium-Run Stock Price Movements?[1]

Stefan Pasch[2]        Daniel Ehnes[3]

## Abstract

To answer this question, we fine-tune transformer-based language models, including BERT, on different sources of company-related text data for a classification task to predict the one-year stock price performance. We use three different types of text data: News articles, blogs, and annual reports. This allows us to analyze to what extent the performance of language models is dependent on the type of the underlying document. StonkBERT, our transformer-based stock performance classifier, shows substantial improvement in predictive accuracy compared to traditional language models. The highest performance was achieved with news articles as text source. Performance simulations indicate that these improvements in classification accuracy also translate into above-average stock market returns.

---

[1] **DISCLAIMER:** The views and opinions expressed in this article are those of the authors and do not reflect any views or positions of any affiliated organization of the authors.
This work is for research purposes **only** and **does not** provide any financial or trading guidance.
[2] Stefan Pasch: stefan.pasch@outlook.com
[3] Daniel Ehnes: danielehnes@gmail.com

# 1. Introduction

In recent years, transformer-based language models have received strong interest in both academic circles and the industry. The development of these language models has accelerated since the publication of BERT (Bidirectional Encoder Representations from Transformers) (Devlin et al. 2018), which set new high scores in various NLP tasks and applications, such as question-answering, fact-checking, or hate-speech detection.

Unsurprisingly, transformer models have influenced the literature and work that links textual information, such as corporate communication or tweets, with stock price performance. While initial attempts to use NLP methods for this purpose have mostly relied on bag-of-word type models (Loughran and McDonald 2011; Loughran and McDonald 2016; Jiang et al. 2019), transformer-based approaches allow to capture more complex aspects of textual information. By integrating contextual information into the evaluation of words, initial evidence suggests that transformer models indeed achieve higher accuracy in predicting stock price movements compared to previous approaches that are, for example, based on naive bayes or dictionary methods (Chen 2021; Sivri et al. 2022).

Most of this work has focused on the stock movements in the immediate aftermath of the publication of the corresponding documents, i.e., 5 to 30 days after a publication event. Yet, many investors do not engage in day-trading and are interested in stock performance over a longer time span. A natural question that arises is whether predictions would also work on longer time horizons. Moreover, most language models that try to predict stock prices analyze only a single source of text data, e.g., one type of corporate communication. However, some types of company-related text data may be more informative than others. We aim to fill these research gaps in the following way:

First, in this paper, we investigate whether language models are able to predict the stock performance for longer time frames. To do this, we analyze the stock price movements on a one-year time frame after the publication of related text data. Second, we compare three different sources of text data and evaluate to what extent the performance of language models is dependent on the type of underlying documents. Specifically, we use news articles, blog posts and annual reports.

To address these questions, we use a financial news dataset (Gennadiy 2020), containing news articles and opinion pieces, as well as official Form-10K annual report filings for 250 firms between 2012 and 2019. Additionally, we gathered information on the stock performance of these companies. We link these data sources to fine-tune transformer models that classify the one-year stock performance based on the corresponding text data.

Analyzing the performance of these language models for our sample data reveals that our language models are able to classify the one-year performance with an accuracy of up to 10 percentage points above the expected accuracy of a random stock movement classifier. Additionally, we find that BERT-based language models outperform traditional language classifiers in all our specifications. Further, the analysis confirms that the performance of these models is highly dependent on the underlying text data, showing a clear ranking, with news data leading to highest, blog articles to the second highest, and company reports leading to the lowest performance. This also provides interesting economic insights, as our results suggests that news articles contain information that are the most "valuable" to an AI. Potentially, blog articles, in their speculative nature, only add noise compared to news articles, whereas the informational content of company reports may be too sparse.

In our supplementary analysis, we investigate to what extent the performance of the news-based model translates into stock return, showing that for our sample and observation period, the recommended picks indeed perform well, compared to the average performance of the entire sample.

## 2. Related Work

Finance, accounting, and economics scholars have long been interested in the interaction of textual information and stock price movements (Cutler et al. 1989; Tetlock 2007; Groß-Klußmann and Hautsch 2011). Most of these studies link news coverage or corporate communications to stock price movements, initially using rather unsophisticated but nonetheless effective bag-of-words methods, such as sentiment analysis based on dictionaries, to investigate these relationships (Loughran and McDonald 2011). Particularly official corporate disclosures have been scrutinized, with studies finding relationships between the readability of documents and stock returns. Additionally, the tone of the documents and the information provided to investors has an effect on stock market returns (Loughran and McDonald 2016). News articles and their relation to stock return have also received a lot of attention: Researchers found evidence suggesting that rising negative sentiment in a firm's news coverage lowers a firm's returns (Ahmad et al. 2016). Media coverage effects may, however, not be persistent, as increased coverage and visibility, for example, can also generate momentum returns that tend to wane in the long run (Hillert et al. 2014). Similarly, social media sentiment has also been linked to stock market returns (Duz Tan and Tas 2021; Sprenger et al. 2014).

In recent years, deep learning based language models made great strides in better understanding textual information, specifically taking the surrounding context into consideration when building their word representations (Chan et al. 2020). Starting with LSTM-based models, such as ELMO (Peters et al. 2018), and with the introduction of transformer-based models, particularly BERT in 2018 (Devlin et al. 2018), deep learning models have strongly improved the ability of NLP models to answer questions, inference textual meaning or summarize text. Moreover, specifically for text classification, BERT models achieve higher accuracy compared to traditional NLP models, such as TF-IDF based models, in various text classification applications (Gonzalez-Carvajal and Garrido-Merchan 2020).

In the domain of finance, transformer-based approaches also outperform traditional NLP approaches, such as dictionary methods, in various tasks, including sentiment analysis, sentence boundary detection, and question-answering for finance-related texts. Moreover, further performance increases could be achieved by pre-training finance-specific language models like FinBERT (Liu et al. 2020).

Not surprisingly, researchers have integrated these deep learning approaches to predict stock price movements by applying such language models on company-related text data, such as tweets or company reports. For example, Sawhney et al. (2020) combine the word encodings in twitter texts with graph neural networks to predict whether stocks decrease or increase within a 5-day lag window. Similarly, Sonkiya et al. (2021) use BERT and GAN to predict stock prices based on news articles in a 5 to 30 days window. However, to the best of our knowledge, there has been no investigation on the ability to predict medium-run stock movements utilizing this technology.

# 3. Data

## 3.1. Company-Related Articles

One of our principal goals is to test what type of textual information is most useful to predict medium-run stock price development. Accordingly, we scrutinize three types of company related text data: First, we utilized a dataset of historical financial news coverage available from

Kaggle.com (Gennadiy 2020). The dataset covers company-specific news articles (henceforth "news") and stock market opinion pieces (which we refer to as "blogs"). Second, we use annual reports as filed with the Securities and Exchange Comminssion (SEC), the so-called Form-10K filings. Together, these sources represent examples of three very important sources of text based information for investors. Annual reports reflect the corporate communications with their stakeholders directly (other examples would be Earnings Calls or Form-8K Filings), news coverage represents how financial news organizations cover changes in the companies' operation and its prospects, while blogs often incorporate professional and semi-professional analysis of stock price movements.

The historical financial news dataset covers 95,578 news articles and 125,935 blogs related to U.S. publicly traded equities covering 800 different corporations (all listed at the NASDAQ or NYSE). The timeframe of the dataset itself starts in October 2008 and continues through February 2020. We made certain restrictions and did not include the entire dataset: First, since the amount of coverage an individual equity receives is highly skewed, for example towards large firms and stocks that have larger trading volumes, we truncated the number of equities to the 250 corporations that received the most coverage in the dataset. This ensures that every equity has at least 170 individual items associated with it. Additionally, the temporal distribution of entries is also highly skewed toward the later years. Therefore, we truncated the dataset to only include 2012 forward in our analysis, as years before then have partially less than 1000 news items in total.

For the annual reports we relied on the data provided by Bill McDonald and Tim Loughran at the University of NotreDame. We relied on all available annual reports from the same companies that we selected from the news and blog dataset. In contrast to the news and blog data, annual reports are only reported once a year, and are overall much more extensive. To deal with this, we split the reports into individual paragraphs.

## 3.2. Sampling

To prepare the data further, we split the data into a train, development, and test set. The train set covers the years 2012-2017 and comprises 225 equities. The dev set covers 10% of the companies (25) and also comprises the years 2012-2017. Our test set covers the same 225 firms from the train set, but in the entire year of 2019.[4] Essentially, there is a one-year gap between the train and the test set. This is necessary because we want to predict the average stock return after one year. Therefore, if we include data from 2018 there would be a large overlap in the stock-price developments of the observations in the training set and the test set, which could lead to spurious results. Accordingly, we did not use the news and blog data from 2018 in our main specification.[5]

---

[4] Please note that since the data for annual reports stops after 2018 we conduct the same splitting rule as for the news- & blog articles with one year lag. This means, the test period includes reports from 2018.

## 3.3. Stock Price Movements

For these companies we also gathered daily stock price data from Yahoo Finance. To abstract from general market movements, we look at the abnormal price in- or decreases for each stock, that is, a stock's price change compared to the average price change of the market. These abnormal one-year return data were then split in tertiles, dividing stocks in three equally large groups of over-, under-, and average-performers. Hence, from a random classifier we would expect an accuracy of 33%.

# 4. Training

## 4.1. Transformer-Based Models

We fine-tuned various transformer-based language models to classify the above-mentioned texts based on the expected stock performance of the corresponding company. We will mainly present results for fine-tuning BERT-Base (Devlin et al. 2018), as this gave us the highest accuracy, but also show results for FinBERT (Araci 2019), and RoBERTa-Large (Liu et al. 2019), BERT-Large (Devlin et al. 2018), and Electra-Large (Clark et al. 2020) in Appendix A1. For all models, we use the following fine-tuning specifications:

**TABLE 1**
**Hyperparameters Transformer-Based Models**

| Hyperparameter | Values |
|---|---|
| Learning Rate | 1e-5 |
| Batch Size | 16 |
| Max Sequence Length | 200 |
| Number of Epochs | 1 |
| Dropout Rate | 0.1 |

---

[5] In a robustness check, we used the 2018 data as our test set achieving comparable results.

Note that in Appendix A2 we also test BERT-Base with a higher number of epochs. If anything, the performance tends to decrease with increasing epochs.

## 4.2. Traditional Text Classifiers

A classical way to conduct supervised NLP-tasks is to consider the input text as a bag-of-words and analyze their frequency of occurrence (Gonzalez-Carvajal and Garrido-Merchan 2020). A common approach to do so is by forming Term Frequency - Inverse Document Frequency (TF-IDF) matrices that measures how often a word occurs in a text relative to the inverse number of occurances in the entire document corpus. Hence, we vectorize the text inputs using *TfidfVectorizer* from *sklearn*, similar to Gonzalez-Carvajal and Garrido-Merchan (2020). Based on these vectorized text inputs, in a second step, traditional machine learning algorithms can be applied to predict the class of the corresponding text inputs. In particular, we will use Logistic Regressions and XGBoost as benchmarks.

# 5. Results

## 5.1. Classifier Results

Table 2 reports the performance of fine-tuned BERT models (StonkBERT) on a hold-out test set for the three text sources. Our first finding is that StonkBERT consistently outperforms the traditional methods in predicting the return category, irrespective of the data source under scrutiny. In fact, in some of the cases, a random draw would have outperformed the traditional models, while the StonkBERT results consistently outperform a random draw, albeit only narrowly for the annual reports.

More importantly, we find that the text source we use has a sizable influence on the achieved accuracy. The lowest performance is achieved using annual reports, indicating that the information therein is largely already priced in or at least of limited long-term value to investors. In contrast, models trained on blogs and news articles performed better. Comparing the accuracy between news and blogs, we find that the informational content in news is most valuable. Our best model achieves an accuracy of 43%, i.e., 10 percentage points over a random draw. Note that this is roughly in line with the work that focuses on short-term stock price movements. For instance, Babbe et al. (2019) outperform a simple one-class classifier by 5 percentage points and Sawhney et al. (2020) outperform a random classifier by 10 percentage points. Training language models on multiple text sources combined did not result in higher accuracy (not shown).

**TABLE 2**
**Classification Results**

| Model/ Data | News | Blogs | Company Reports |
|---|---|---|---|
| Random | Acc.: 0.33 | Acc.: 0.33 | Acc.: 0.33 |
|  | F1: 0.33 | F1: 0.33 | F1: 0.33 |
| Logistic Regression | Acc.: 0.38 | Acc.: 0.36 | Acc.: 0.33 |
|  | F1: 0.36 | F1: 0.35 | F1: 0.33 |
| XGBoost | Acc.: 0.33 | Acc.: 0.33 | Acc.: 0.24 |
|  | F1: 0.26 | F1: 0.27 | F1: 0.17 |
| **StonkBERT** | **Acc.: 0.43** | **Acc.: 0.39** | **Acc.: 0.36** |
|  | **F1: 0.43** | **F1: 0.39** | **F1: 0.37** |

## 5.2. Performance Analysis

Though the language models were trained on a simple classification task with three categories, we also analyze in how far the model's predictions translate into stock market returns. This means, we analyze the average abnormal one-year return in our news article test sample based on the predictions from StonkBERT. First, we calculate the performance of the three prediction groups (good, medium, bad) in our test period, where performance is measured as the average abnormal one-year performance in a rolling window.[6] Table 3 reports the results. In our test set, the included companies showed an average performance of 6.29%. The firms that had been predicted as "good" by StonkBERT, however, showed an average performance of 16.83% in the one-year period after the predictions were made. The firms that were predicted as "average" just showed a performance of 4.72% and the "bad" predictions of -3.17%, respectively. Correspondingly, the classification into the three performance groups was actually associated with substantial differences in the one-year stock returns. Moreover, we looked at the Top-10 predictions, which include the 10 firms where StonkBERT predicted the highest probability of being a "good" firm. Those firms in fact outperformed the market by an even higher margin with an average one-year performance of over 41%. For the Flop-10 predictions, however, we do not find that they further underperformed compared to all firms that were predicted as "bad".

---

[6] We calculated the one-year return for every trading day, over the entire year and then calculated the average over all days.

We also conducted performance simulations for articles published in year 2018 and also find performance differences between the predicted categories, albeit the outperformance of the Top-10 compared to the market appeared to be more modest.

TABLE 3
Performance Simulations

| Grouping | Year 2019 (Test Set) | Year 2018 (Robustness Check) |
|---|---|---|
| Whole Sample | 6.29% | 5.24% |
| Prediction: Good | 16.83% | 12.33% |
| Prediction: Average | 4.72% | 6.40% |
| Prediction: Bad | -3.17% | -5.49% |
| Prediction: Top-10 | 41.02% | 8.65% |
| Prediction: Flop-10 | -1.85% | -0.75% |

The following figures show the average stock price development for the different groups comprising the entire two-year period, with analyzed articles covering the entire year of 2019 only. Interestingly, we find that the firms predicted as "good" outperformed the other categories in both time periods before and after the Covid related stock market crash in March 2020.

FIGURE 1
Performance Simulations over Time

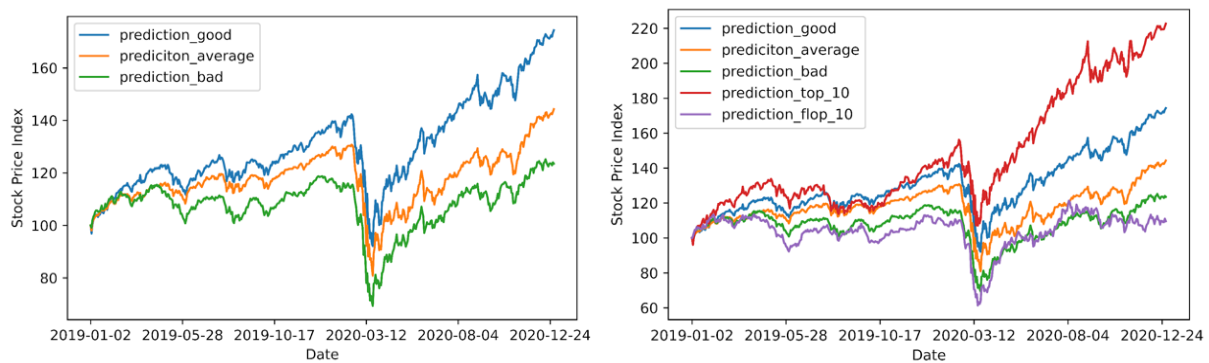

# 6. Discussion

We set out to test whether transformer-based language models can learn valuable information from text data to predict the stock performance of affiliated firms over a one-year time-period. Our results provide two interesting findings: First, the predictive capability of the model heavily depends on the informational value of the underlying text data, where in all specifications and models the text data from the news sample outperformed both the blog as well as the annual reports sample. Further, we found that state-of-the-art transformer-models indeed outperform traditional NLP approaches to predict stock returns. While our results are encouraging, there are several important limitations that need to be kept in mind.

## 6.1. Economic Explanation

Among finance scholars the dominant theory is the efficient market hypothesis (Fama 1970), which suggests that stock prices reflect all available information of the market, making it impossible to systematically outperform the market. Correspondingly, our results should be taken with a grain of salt and there exist various explanations why NLP models including StonkBERT may not outperform the market in the long run.

First, one explanation for the successful prediction may be the result of our specific time period, which includes very strong outperformance of tech-based stocks in general that were further amplified after the Covid crisis emerged. Correspondingly, our model may have learned that tech-based stocks outperformed their peers during the training period and inferred this trend to continue in the future. However, a purely industry-based effect should have been detected by traditional models as well. Potentially, transformer-based models may be able to pick up more fine-grained information, for example specific technological trends within industries (e.g. cloud computing, or machine learning). Another reason why we suspect that the results are not entirely driven by industry based effects is that the annual report based text data was unable to pick up on such industry based effects, despite convincing and thorough evidence that their contents are an excellent predictor of a firm's industry (Hoberg and Phillips 2016).

Similarly, our models may mainly capture a momentum-effect that has been widely studied in the finance literature (Jegadeesh and Titman 2001). Moreover, as our models learn from historical success factors it could also be susceptible to run into stock bubbles. However, it should be noted that the StonkBERT model not just predicted outperformers, which potentially reflect a bubble, but was also able to detect underperformers.

## 6.2. Differences Between Different Text Data

We find the highest accuracy for language models working with news data, the second highest with blog articles, and the worst accuracy for annual reports as text source. A potential explanation for the comparably weak results for our annual report models could be that the information density in annual reports are too sparse, because, for example, they contain various standard phrases and largely contain information on past events that could be already priced in by the market (Yuan et al. 2021). Another concern is the limited frequency, which means that only new information with a close temporal proximity to the report can be learned by the model. Further, seasonal fluctuations are also not included, as the annual reports are generally released around the same time each year. An interesting avenue of future research could therefore be to test other more frequent corporate communication such as earnings reports, quarterly reports or form-8K filings, which could disentangle the frequency concerns from the considerations around limited informational content in such documents.

For the performance difference between blogs and news, a potential explanation in our estimation is the increased noise that is created through speculative attempts by bloggers to beat the market. Additionally, since these articles usually are framed as opinion pieces, their average news content should be lower compared to news articles.

## 6.3. Technical Considerations

For most language tasks, larger transformer models tend to outperform smaller models, e.g. BERT-Large outperforms BERT-Base in the seminal BERT paper (Devlin et al. 2018). We, however, find that no model, including various "large" models could outperform a BERT-Base model. Moreover, we achieved the best results with just one epoch of fine-tuning the language model for the stock performance classification task. A potential explanation could be that the classification task at hand is susceptible to overfitting.

Another open question is the length of the training period and test period. In our approach, we used a simple heuristic based on the amount of data available. For example, including news data from the 1980s is unlikely to improve the predictive performance of the model for today, it may even introduce noise or outdated information that decrease the models' performance. On the other hand, using a longer training period could prevent an overfitting on short term trends that do not reflect fundamental values. Similarly, for the test period, it is so far unclear how far the performance differences between predicted groups persist after the one-year period.

# Publication bibliography

# Appendix A1

**TABLE A1**
**Comparing Transformer Models (News Articles)**

| Model/ Data | Performance |
|---|---|
| **BERT-base (StonkBERT)** | **Acc.: 0.43** <br> **F1: 0.43** |
| BERT-large | Acc.: 0.42 <br> **F1: 0.43** |
| FinBERT | Acc.: 0.42 <br> **F1: 0.43** |
| **RoBERTa-large** | **Acc.: 0.43** <br> **F1: 0.43** |
| Electra-large | Acc.: 0.39 <br> F1: 0.38 |

# Appendix A2

**TABLE A2**
**StonkBERT Sensitivity Analysis Number of Epochs**

| Number of Epochs | Performance on Newsdata |
|---|---|
| **Number of Epochs = 1** | **Acc.: 0.43** <br> **F1: 0.43** |
| Number of Epochs = 2 | Acc.: 0.40 <br> F1: 0.40 |
| Number of Epochs = 3 | Acc.: 0.42 <br> F1: 0.41 |
| Number of Epochs = 4 | Acc.: 0.41 <br> F1: 0.40 |